\def\paperlanguage{}
\pdfoutput=1 

\documentclass[letterpaper, 10 pt, conference]{ieeeconf}  

\usepackage{bm}
\usepackage{cite}
\include{preamble}

\newcommand{\ctext}[1]{\raise0.2ex\hbox{\textcircled{\scriptsize{#1}}}}

\IEEEoverridecommandlockouts                              

\title{\LARGE \textbf
  {
    \switchlanguage%
    {%
      Robot Design Optimization with Rotational and Prismatic Joints\\ using Black-Box Multi-Objective Optimization
    }%
    {%
      回転関節と直動関節を含むロボット設計の多目的最適化
    }%
  }
}

\author{Kento Kawaharazuka$^{1}$, Kei Okada$^{1}$, and Masayuki Inaba$^{1}$
  \thanks{$^{1}$ The authors are with the Department of Mechano-Informatics, Graduate School of Information Science and Technology, The University of Tokyo, 7-3-1 Hongo, Bunkyo-ku, Tokyo, 113-8656, Japan.
    {\texttt\small [kawaharazuka, k-okada, inaba]@jsk.t.u-tokyo.ac.jp}
  }
}
\begin{document}

\maketitle
\thispagestyle{empty}
\pagestyle{empty}

\begin{abstract}
  \switchlanguage%
  {%
    Robots generally have a structure that combines rotational joints and links in a serial fashion.
    On the other hand, various joint mechanisms are being utilized in practice, such as prismatic joints, closed links, and wire-driven systems.
    Previous research have focused on individual mechanisms, proposing methods to design robots capable of achieving given tasks by optimizing the length of links and the arrangement of the joints.
    In this study, we propose a method for the design optimization of robots that combine different types of joints, specifically rotational and prismatic joints.
    The objective is to automatically generate a robot that minimizes the number of joints and link lengths while accomplishing a desired task, by utilizing a black-box multi-objective optimization approach.
    This enables the simultaneous observation of a diverse range of body designs through the obtained Pareto solutions.
    Our findings confirm the emergence of practical and known combinations of rotational and prismatic joints, as well as the discovery of novel joint combinations.
  }%
  {%
    一般的にロボットは, 回転関節とリンクをシリアルに組み合わせた構造を持つ.
    一方で, 直動関節や閉リンク, ワイヤ駆動など, 実際には多くの関節機構が存在している.
    これまでの研究では, それら一つ一つの機構に着目し, その配置やリンクの長さを最適化することで, 与えられたタスクを達成可能なロボットの設計手法が提案されてきた.
    そこで本研究では, 回転関節と直動関節という異なる関節を組み合わせたロボットの設計最適化を行う手法を提案する.
    関節数とリンク長を最小化しつつ, 目的のタスクを達成するロボットを, ブラックボックスな多目的最適化に基づき自動生成する.
    得られたパレート解から多様な身体設計を一度に観察することができる.
    既存の考えられてきた実用的な関節機構の組み合わせ, 新たな関節機構の組み合わせ等が発現することを確認した.
  }%
\end{abstract}

\section{INTRODUCTION}\label{sec:introduction}
\switchlanguage%
{%
  Robots typically feature a structure that combines rotational joints and links in a serial manner \cite{willow2010pr2, kaneko2004hrp2}.
  On the other hand, in practice, there exist various joint mechanisms such as prismatic joints \cite{urata2016leg, wang2020slider}, closed links \cite{siciliano1999closed}, and wire-driven systems \cite{kawaharazuka2019musashi, yoshimura2023arrangement}.
  Historically, these designs have been manually crafted by humans, but there are numerous initiatives aimed at automating these processes.

  \cite{yang2000modular} optimized the number and types of one-axis rotational joint modules, as well as their relative positions, for industrial robots using a genetic algorithm to achieve desired operational points.
  \cite{xiao2016designopt} optimized the motors and gear ratios of a general serial-link 6-DOF manipulator based on weight minimization and manipulability maximization, though it does not involve modular robots.
  \cite{kawaharazuka2023autodesign} performed multi-objective optimization for a modular robot composed of one-axis rotational joints, focusing on minimizing positional error and force for daily life support.
  In the realm of closed-link structures, \cite{kim2003parallelopt} optimized the link lengths and configurations.
  For wire-driven systems, \cite{hamida2021arrangement} optimized the wire attachment positions for a parallel wire-driven robot using evolutionary computation.
  \cite{kawaharazuka2024arrangeopt} extended \cite{hamida2021arrangement} to include wire attachment links under selectable conditions, enabling more complex optimization.
}%
{%
  一般的なロボットは, 回転関節とリンクをシリアルに組み合わせた構造を持つ\cite{willow2010pr2, kaneko2004hrp2}.
  その一方で, 直動関節\cite{urata2016leg, wang2020slider}や閉リンク\cite{siciliano1999closed}, ワイヤ駆動\cite{kawaharazuka2019musashi, yoshimura2023arrangement}など, 実際には多くの関節機構が存在している.
  これまで, それらの設計は人間が一つ一つ自ら考えて設計を行ってきたが, それらを自動化しようという取り組みも多数存在している.
  \cite{yang2000modular}は産業用ロボットに向け, 所望の動作点を満たす1軸の回転関節モジュールの数と種類, モジュール間の相対的な位置を遺伝的アルゴリズムにより最適化した.
  \cite{xiao2016designopt}はモジュールではないが, 重量の最小化とmanipulabilityの最大化に基づき一般的なシリアルリンク6自由度マニピュレータのモータとギア比を最適化した.
  \cite{kawaharazuka2023autodesign}は日常生活支援に向け, 位置誤差と力の最小化に基づく1軸の回転関節により構成されるモジュラーロボットの多目的最適化を実行した.
  \cite{kim2003parallelopt}は閉リンク構造におけるリンク長や配置の最適化を行った.
  \cite{hamida2021arrangement}はCable-driven parallel wire robotについて, そのワイヤ取り付け位置を進化計算に基づき最適化している
  \cite{kawaharazuka2024arrangeopt}はワイヤの取り付け位置だけでなく, ワイヤの取り付けリンクも選択可能な条件で, より複雑な最適化を実行している.
}%

\begin{figure}[t]
  \centering
  \includegraphics[width=0.95\columnwidth]{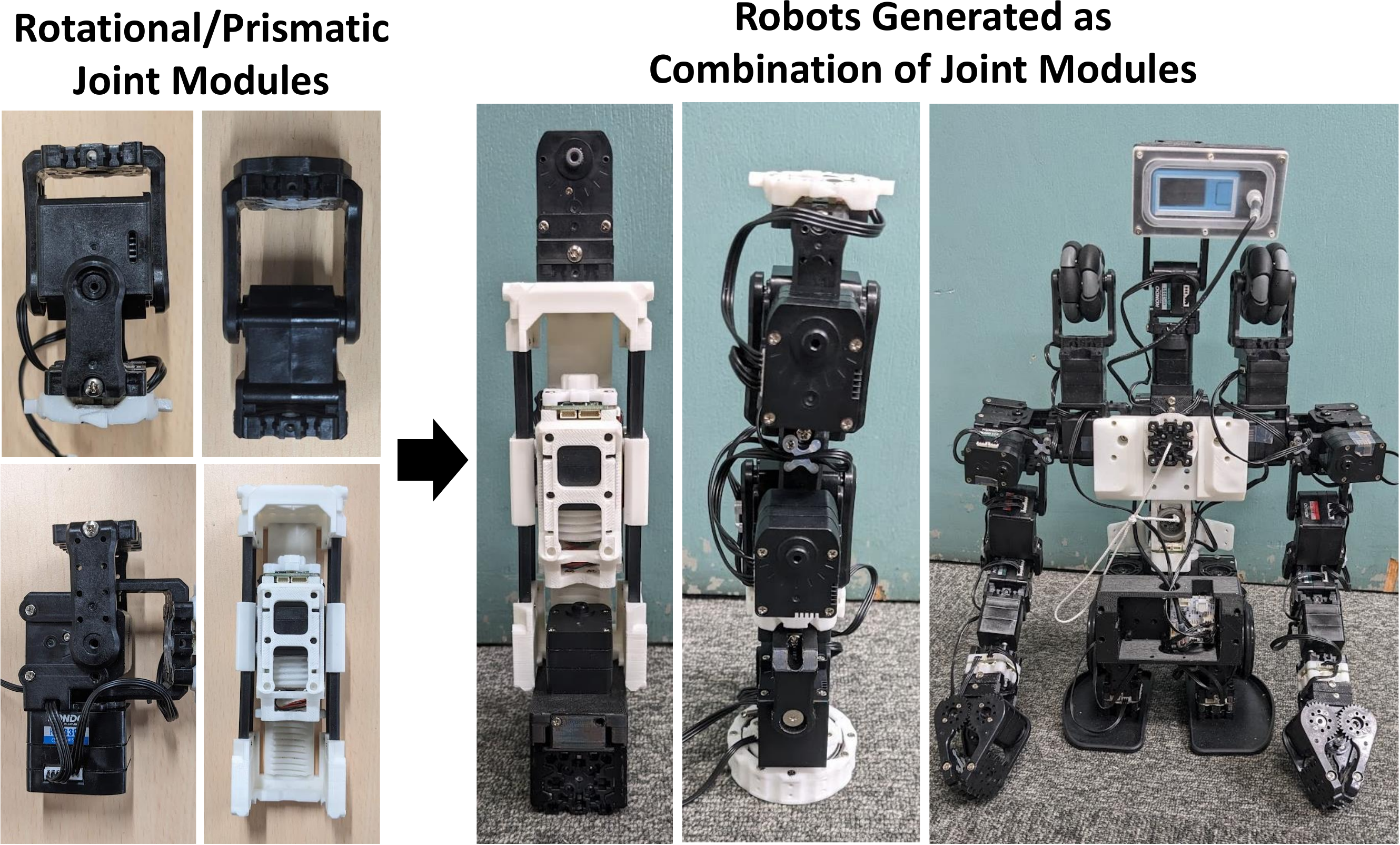}
  \vspace{-1.0ex}
  \caption{Examples of rotational and prismatic joint modules, as well as robots generated through their combinations.}
  \label{figure:example}
  \vspace{-1.0ex}
\end{figure}

\begin{figure*}[t]
  \centering
  \includegraphics[width=1.9\columnwidth]{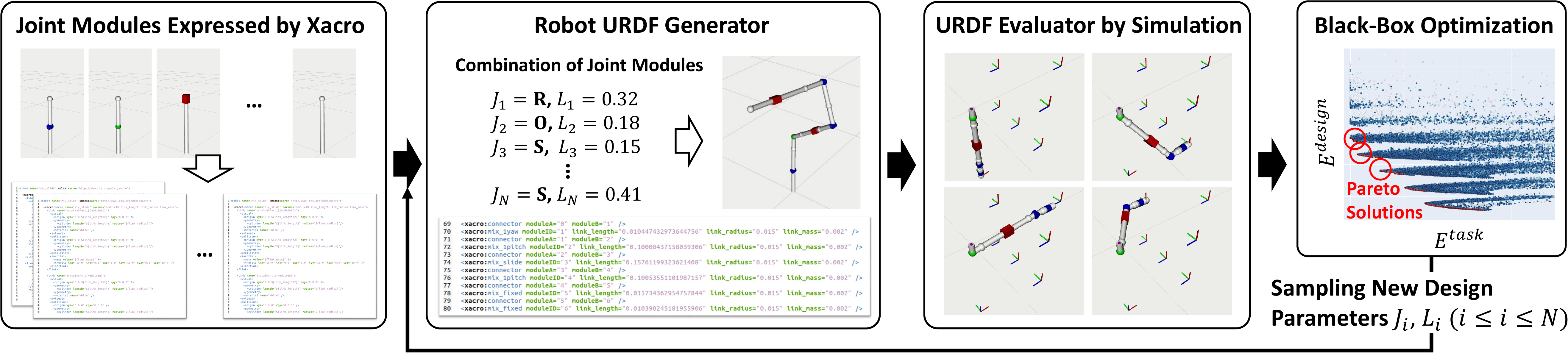}
  \vspace{-1.0ex}
  \caption{The overview of the proposed system. Joint modules are expressed using Xacro, and a robot URDF is generated by combining these modules. The generated URDF is then evaluated through simulation, and the subsequent designs are iteratively generated by black-box optimization.}
  \label{figure:system}
\end{figure*}

\begin{figure*}[t]
  \centering
  \includegraphics[width=1.9\columnwidth]{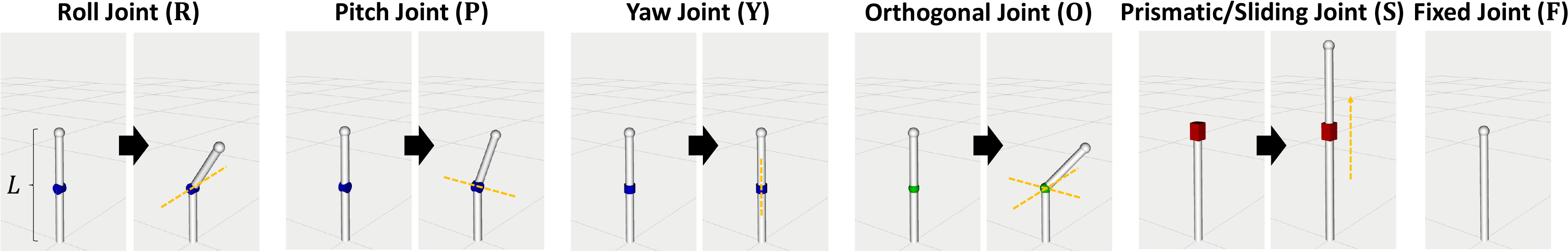}
  \vspace{-1.0ex}
  \caption{The joint modules used in this study. The roll joint (\textbf{R}), the pitch joint (\textbf{P}), the yaw joint (\textbf{Y}), the two-axis orthogonal joint (\textbf{O}), the prismatic/sliding joint (\textbf{S}), and the fixed joint (\textbf{F}) are predefined. The design parameters encompass the type of the joint module and the length of the link $L$.}
  \label{figure:modules}
  \vspace{-1.0ex}
\end{figure*}

\switchlanguage%
{%
  Despite various research efforts in design optimization, the focus has predominantly been on optimizing individual mechanical structures.
  Notably, there have been several recent studies specifically addressing design optimization of bodies incorporating one-axis rotational joints and rotational wheels.
  \cite{zhao2020robogrammar, xu2021multi} optimized the joint arrangement and discrete values of link lengths for a modular robot capable of traversing uneven terrain.
  Similarly, \cite{hu2022modular} conducted design optimization for a modular robot adept at navigating rough terrain, utilizing Generative Adversarial Network (GAN) for diverse body generation.

  On the other hand, despite the existence of numerous combinations of various joint mechanisms, a considerable portion of them remain unexplored.
  More specifically, examples that combine prismatic joints, closed-link structures, and wire-driven systems are not apparent, leaving untapped potential for numerous valuable solutions.
  \figref{figure:example} illustrates such examples, showcasing the possibility of constructing diverse robot bodies through various combinations of joint structures.
  These constructions should not be limited to human-designed solutions but should be autonomously discovered.
  Therefore, in this study, we conduct design optimization for robots by combining multiple joint structures, particularly incorporating both rotational and prismatic joints.
  Each joint structure is modularized and described as an individual file, forming the robot model through their combinations.
  The objective is to minimize the number of joints and link lengths while achieving the given task, employing a black-box multi-objective optimization for automatic design optimization.
  This research stands out from previous studies by not only focusing on robot performance but also emphasizing design aspects such as minimizing the number of joints and link lengths.
  Additionally, by intentionally avoiding a single objective function and employing multi-objective optimization, we can obtain a multitude of solutions with varying degrees of freedom as Pareto solutions, allowing us to grasp the characteristics of each design.
  From the obtained Pareto solutions, we confirmed the emergence of practical and known combinations of rotational and prismatic joints, as well as the discovery of novel joint combinations.

  The structure of this study is outlined as follows.
  In \secref{sec:proposed}, we elaborate on design parameters, the definition of objective functions, and multi-objective optimization.
  In \secref{sec:experiment}, we apply the proposed method to optimize robot design for achieving three specific tasks.
  Finally, in \secref{sec:conclusion}, we present conclusions and future prospects.
}%
{%
  このように様々な設計最適化研究が行われてきた一方で, それらはどれも一つ一つの機械構造に着目した設計最適化を行っており, それらの配置やリンク長等を最適化している.
  その中でも, 近年一軸の回転関節と回転車輪を含む身体設計最適化の研究がいくつか行われている.
  \cite{zhao2020robogrammar}は不整地を走行可能なモジュラーロボットの関節配置やリンク長の離散値に関する最適化を行った.
  \cite{hu2022modular}はGenerative Adversarial Network (GAN)を用いた多様な身体生成と不整地走行可能なモジュラーロボットの設計最適化を行った.

  一方で, その他多様な組み合わせがあるにも関わらず, それらは探索されていない.
  特に直動関節やワイヤ駆動, 閉リンクを組み合わせた例は見当たらず, 有益な解が多数残されている可能性がある.
  \figref{figure:example}にはその例を示すが, 様々な関節構造の組み合わせにより, 多様な身体が構成可能であり, それらは人間の手ではなく, 自律的に発見されるべきである.
  そこで本研究では, 複数の関節構造, 特に回転関節と直動関節を組み合わせたロボットの設計最適化を行う.
  各関節構造を個別のファイルとしてモジュール化して記述し, それらの組み合わせとしてロボットモデルを構成する.
  この際, 関節数とリンク長の最小化, 与えられたタスクの達成を目的関数とし, ブラックボックスな多目的最適化に基づき自動的な設計最適化を行う.
  本研究はこれまでの研究とは異なり, ロボットのパフォーマンスだけでなく, 関節数やリンク長の最小化といった設計上の観点に着目していることも重要な要素である.
  また, 敢えて単一の目的関数は作らずに多目的最適化を行うことで, 自由度数の異なる多数の解をパレート解として一度に得ることができ, それぞれの設計の特徴を把握することができる.
  得られたパレート解から, 既存の実践的な関節機構の組み合わせ方に加え, 新たな関節機構の組み合わせが発現することを確認した.

  本研究の構成は以下である.
  \secref{sec:proposed}では, 身体パラメータの構築と評価関数の定義, 多目的最適化について述べる.
  \secref{sec:experiment}では, 本研究で提案する手法を用いて, 3つのタスクを達成するロボットの設計最適化を行った.
  \secref{sec:conclusion}で結論と今後の展望を述べる.
}%

\section{Robot Design Optimization with Rotational and Prismatic Joints using Black-Box Multi-Objective Optimization} \label{sec:proposed}
\switchlanguage%
{%
  The overall system is illustrated in \figref{figure:system}.
  Each joint module is described using Xacro (XML Macros), and these modules are combined to generate the URDF (Unified Robot Description Format) of the robot.
  This URDF is then evaluated through simulation, and new design parameters are iteratively generated using black-box optimization.
  In this section, we first discuss the structure of each joint module, how they interconnect to form a unified robot, and describe the set of body parameters.
  Next, we define objective functions to evaluate these parameters and describe black-box multi-objective optimization based on these evaluations.
  }%
  {%
  全体のシステム構成を\figref{figure:system}に示す.
  各関節モジュールはXacro (XML Macros)で記述され, これらを組み合わせてロボットのURDF (Unified Robot Description Format)を生成する.
  これをシミュレーションで評価し, ブラックボックス最適化により繰り返し新しい設計パラメータを生成する.
  本章ではまず, どのように各モジュールを構成し, それらをどう繋げて一つのロボットとするのか, 設定した身体パラメータについて述べる.
  次に, それらを評価する関数を定義し, これに基づくブラックボックス多目的最適化について述べる.
}%

\subsection{Design Parameters} \label{subsec:design-params}
\begin{figure}[t]
  \centering
  \includegraphics[width=0.95\columnwidth]{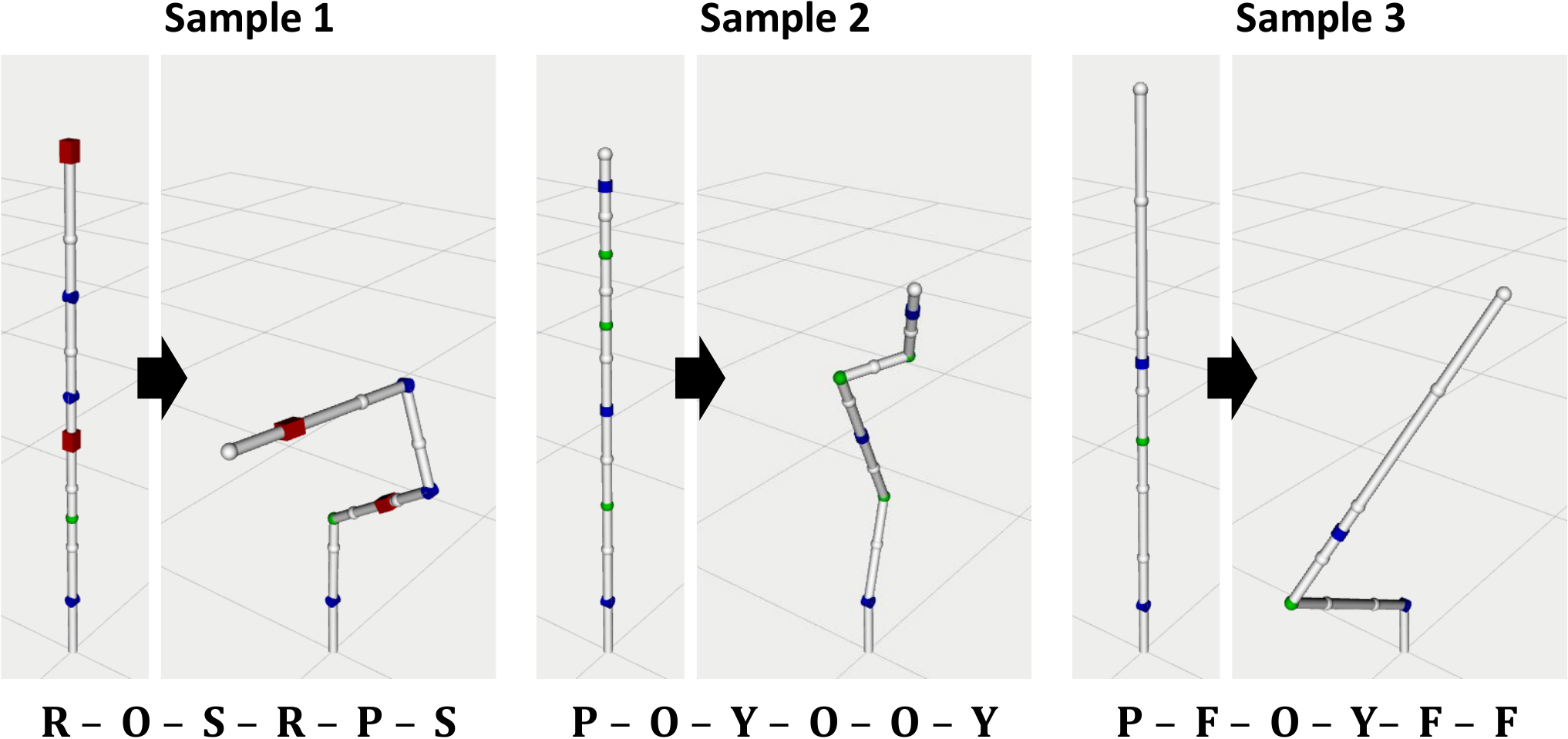}
  \caption{The random robot models generated through the combination of predefined joint modules.}
  \label{figure:random}
\end{figure}

\switchlanguage%
{%
  We first describe the prerequisites for the body design.
  We denote the position and orientation of the robot's root link as $\bm{p}^{root}$ and $\bm{R}^{root}$, respectively.
  Similarly, the position and orientation of the effector are represented as $\bm{p}^{ee}$ and $\bm{R}^{ee}$.
  In the initial pose, where all joint angles and positions are set to 0, we configure the links in a straight line.
  This configuration is similar to that of industrial robots and humanoids; for example, the arms and legs of a humanoid would be straightened in the initial pose.
  While alternative configurations are possible, we adopt this general assumption here.

  Next, we elaborate on each module constituting the robot.
  This study focuses on rotational joints and prismatic joints, with particular emphasis on various configurations, especially in the case of rotational joints.
  The one-axis rotational joints for roll \textbf{R}, pitch \textbf{P}, and yaw \textbf{Y}, as well as the commonly used orthogonal two-axis joint \textbf{O} for roll and pitch, are considered.
  Adding a prismatic joint \textbf{S}, we illustrate the five joints in \figref{figure:modules}.
  Each module has a parameter $L$ representing its length, indicating the total length of the module in the initial pose.
  Rotational joints are positioned between two links of length $L/2$.
  Prismatic joints start with an initial state featuring a link of length $L$ and can extend up to $2L$.
  Additionally, each joint module has a range of motion denoted as $\theta^{\{min, max\}}$.
  In this study, considering general range of motion limits, we set $\theta^{\{min, max\}}=\{-3/4\pi, 3/4\pi\}$ for roll and pitch (including orthogonal two-axis joints), $\theta^{\{min, max\}}=\{-2\pi, 2\pi\}$ for yaw, and $\theta^{\{min, max\}}=\{0, L\}$ for prismatic joints.
  However, for optimization, the number of parameters to optimize must be predetermined.
  Therefore, to avoid initially fixing the number of joints in the robot to be designed, we introduce an additional joint, a fixed joint \textbf{F}.
  This joint is also represented as a link of length $L$, with no articulated joint, and $\theta^{\{min, max\}}=\{0, 0\}$.
  These six modules are described in Xacro format.

  Finally, we combine these modules to construct a robot model.
  By connecting the distal link of the $i$-th module with the proximal link of the $(i+1)$-th module, the links are joined.
  Therefore, with the number of modules included in a single robot denoted as $N^{jnt}$, the design parameters consist of six discrete parameters $J_i$ ($1 \leq i \leq N^{jnt}$) representing the choices of the six joint types: \{\textbf{R}, \textbf{P}, \textbf{Y}, \textbf{O}, \textbf{S}, \textbf{F}\}, and continuous parameters $L_i$ representing the link lengths.
  Here, careful consideration is needed for the range $[L^{min}, L^{max}]$ of $L_i$.
  When $L_i$ is very small, it becomes challenging to construct actual modules especially for roll, pitch, and prismatic joints.
  Therefore, in this study, the range of $L_i$ for \textbf{R}, \textbf{P}, \textbf{O}, \textbf{S} is set as $[0.1, 0.5]$, while for \textbf{Y} and \textbf{F}, it is set as $[0.01, 0.5]$.
  During actual optimization, continuous parameters $C_i$ with a range of $[0.0, 1.0]$ are optimized, and we set $L_i = (L^{max} - L^{min})C_i + L^{min}$.

  The robots constructed with random parameters are shown in \figref{figure:random}.
  It can be observed that diverse body structures are represented by the parameters of the six module types and their corresponding lengths.
}%
{%
  まず身体設計の前提条件について述べる.
  本研究のロボットのroot linkの位置姿勢を$\bm{p}^{root}$, $\bm{R}^{root}$, エンドエフェクタの位置姿勢を$\bm{p}^{ee}$, $\bm{R}^{ee}$とする.
  ここから, 初期姿勢(全ての関節角度や関節位置が$0$となる姿勢)において, 一直線にリンクが並ぶような配置を行う.
  これは産業用ロボットやヒューマノイドと同様の構成であり, 例えばヒューマノイドについて, 腕や脚が初期姿勢で真っ直ぐ伸びたような形になることを意味する.
  もちろん他の構成も考えられるが, ここではこの一般的な仮定を採用した.

  次にロボットを構成する各モジュールについて述べる.
  本研究では回転関節と直動関節について扱うが, 特に回転関節は様々な形が考えられる.
  ロール関節\textbf{R}, ピッチ関節\textbf{P}, ヨー関節\textbf{Y}の一軸関節, そして, よく用いられるのがロールピッチの直交二軸関節\textbf{O}である.
  これに直動関節\textbf{S}を加えた5つの関節を\figref{figure:modules}に示す.
  各モジュールは長さ$L$のパラメータを持ち, これは初期姿勢におけるモジュールの全長を表す.
  各回転関節は長さ$L/2$の2つのリンクによって挟まれる.
  直動関節は長さ$L$のリンクの状態を初期状態とし, これが$2L$まで伸びることができる.
  また, 各モジュールの関節は可動域$\theta^{\{min, max\}}$を持つ.
  本研究では, 一般的な可動域限界を考慮し, ロールとピッチ(直交二軸関節を含む)について$\theta^{\{min, max\}}=\{-3/4\pi, 3/4\pi\}$, ヨーについて$\theta^{\{min, max\}}=\{-2\pi, 2\pi\}$, 直動関節について$\theta^{\{min, max\}}=\{0, L\}$としている.
  一方で, 最適化をするためには, 最初から最適化するパラメータ数が決まっていなければならない.
  そのため, このままでは最初に設計するロボットの関節数が決まってしまう.
  そこで, もう一つの関節として, 固定関節\textbf{F}を導入する.
  これは同様に長さ$L$のリンクであるが, 関節は存在せず, $\theta^{\{min, max\}}=\{0, 0\}$である.
  これら6つのモジュールをそれぞれXacroフォーマットで記述する.

  最後に, これらのモジュールを組み合わせてロボットモデルを構築する.
  $i$番目のモジュールのdistal linkと$i+1$番目のモジュールのproximal linkを結合することで, リンクを連結する.
  よって, 一つのロボットに含まれるモジュール数を$N^{jnt}$として, 設計パラメータは6つの関節のタイプの選択肢\{\textbf{R}, \textbf{P}, \textbf{Y}, \textbf{O}, \textbf{S}, \textbf{F}\}を持つ離散パラメータ$J_i$ ($1 \leq i \leq N^{jnt}$), リンク長を表現する連続パラメータ$L_i$となる.
  ここで, $L_i$の範囲$[L^{min}, L^{max}]$については工夫を要する.
  $L_i$が非常に小さい場合, 特にロールやピッチ, スライド関節の実際のモジュールを構成することは難しい.
  そこで本研究では, \textbf{R}, \textbf{P}, \textbf{O}, \textbf{S}については$L_i$の範囲を$[0.1, 0.5]$とし, \textbf{Y}と\textbf{F}については$[0.01, 0.5]$としている.
  なお, 実際の最適化の際には, $[0.0, 1.0]$の範囲を持つ連続パラメータ$C_i$を最適化しており, $L_i = (L^{max} - L^{min})C_i + L^{min}$としている.

  ランダムなパラメータによって構成されるロボットを\figref{figure:random}に示す.
  6種類のモジュールとその長さのパラメータによって, 多様な身体構造を表現できていることがわかる.
}%

\subsection{Objective Functions and Black-Box Multi-Objective Optimization} \label{subsec:obj-opt}
\switchlanguage%
{%
  First, we discuss the objective functions to be optimized.
  Since it is a multi-objective optimization, multiple objective functions can be specified.
  In this study, two objective functions are used from the perspective of visibility.

  The first one is the task accomplishment of the robot, denoted by the objective function $E^{task}$.
  It is defined as the minimization of whether the specified $N^{ref}$ positions and orientations ($\bm{p}^{ref}_{i}$, $\bm{R}^{ref}_{i}$) ($1 \leq i \leq N^{ref}$) are achieved by the robot's end effector, as follows,
  \begin{align}
    \bm{q}_{i} &= \text{IK}(\bm{p}^{ref}_{i}, \bm{R}^{ref}_{i}) \\
    E^{task}_{i} &= ||\bm{e}(\bm{q}_{i})||_{2} \\
    E^{task} &= \sum_{i=1}^{N^{ref}} E^{task}_{i}
  \end{align}
  where $\bm{q}_{i}$ represents the joint angles obtained by solving the inverse kinematics $\textbf{IK}$, $\bm{e}(\bm{q}_{i})$ is the error vector (the upper 3 components represent the positional error and the lower 3 components represent the rotational error), and $||\cdot||_{2}$ denotes the L2 norm.
  In other words, the sum of errors in inverse kinematics constitutes $E^{task}$.
  While more complex tasks or dynamic movements, such as walking, can be considered, this study focuses on a configuration that specifically examines the simplest kinematics.

  The second objective function is the minimization of the number of joints and link lengths, defined as $E^{design}$.
  It is defined as follows,
  \begin{align}
    E^{design}_{joint} &= N^{jnt}_{R} + N^{jnt}_{P} + N^{jnt}_{Y} + 2N^{jnt}_{O} + N^{jnt}_{S} \\
    E^{design}_{length} &= \sum_{i=1}^{N^{jnt}} L_i \\
    E^{design} &= E^{design}_{joint} + E^{design}_{length}
  \end{align}
  where $N^{jnt}_{\{R, P, Y, O, S\}}$ represents the number of occurrences of \{\textbf{R}, \textbf{P}, \textbf{Y}, \textbf{O}, \textbf{S}\}.

  Next, we delve into multi-objective optimization.
  In typical optimization scenarios, weights are applied to multiple objective functions, and they are transformed into a single objective function for optimization.
  However, this study aims to discover new body morphologies by generating diverse solutions with inherent trade-offs.
  Therefore, intentionally avoiding narrowing down the solutions to a single one, multi-objective optimization is performed based on the two objective functions, presenting a diverse set of Pareto solutions.
  For this purpose, we employ the NSGA-II algorithm \cite{deb2002nsgaii} from the Optuna library \cite{akiba2019optuna} for black-box optimization.
  The number of trials varies slightly for each task, being around 70,000, and the execution time is approximately 5 hours.
}%
{%
  まず, 最適化する目的関数について述べる.
  多目的最適化であるため, 複数の目的関数を指定することができるが, 本研究では視認性の観点から2つの目的関数を用いる.
  1つ目は, ロボットのタスク達成度であり, 指定した$N^{ref}$個の位置姿勢を満たせるかどうかについて, 以下のように目的関数$E^{task}$を定義し, これを最小化する.
  \begin{align}
    \bm{q}_{i} &= \text{IK}(\bm{p}^{ref}_{i}, \bm{R}^{ref}_{i}) \\
    E^{task}_{i} &= ||\bm{e}(\bm{q}_{i})||_{2} \\
    E^{task} &= \sum_{i=1}^{N^{ref}} E^{task}_{i}
  \end{align}
  ここで, ($\bm{p}^{ref}_{i}$, $\bm{R}^{ref}_{i}$) ($1 \leq i \leq N^{ref}$)はロボットのエンドエフェクタが満たすべき位置姿勢の指令値, $\bm{q}_{i}$は逆運動学$\textbf{IK}$を解いた際の関節角度, $\bm{e}(\bm{q}_{i})$は誤差ベクトル(上3成分は位置誤差, 下3成分は回転誤差を表す), $||\cdot||_{2}$はL2ノルムを表す.
  つまり, 逆運動学の誤差の総和が$E^{task}$となる.
  もちろん, より複雑なタスクや動的な歩行等を考えることもできるが, 本研究の焦点からは外れるため, 最もシンプルなキネマティクスのみに着目した構成としている.
  2つ目は, 関節数とリンク長さの最小化であり, 以下のように目的関数$E^{design}$を定義する.
  \begin{align}
    E^{design}_{joint} &= N^{jnt}_{R} + N^{jnt}_{P} + N^{jnt}_{Y} + 2N^{jnt}_{O} + N^{jnt}_{S} \\
    E^{design}_{length} &= \sum_{i=1}^{N^{jnt}} L_i \\
    E^{design} &= E^{design}_{joint} + E^{design}_{length}
  \end{align}
  ここで, $N^{jnt}_{\{R, P, Y, O, S\}}$はそれぞれ, \{\textbf{R}, \textbf{P}, \textbf{Y}, \textbf{O}, \textbf{S}\}の含まれる数を表す.

  次に, 多目的最適化について述べる.
  大抵の最適化では, 複数の目的関数に重み付けを施し, これを単一の目的関数に変換して最適化を行う.
  一方で本研究は, トレードオフのある多様な解を出力することで, 新しい身体形態を発見することが主な目的である.
  そのため, 敢えて解を一つに絞らず, 2つの目的関数から多目的最適化を行い, その多様なパレート解を提示する.
  本研究ではBlack-Box OptimizationのライブラリとしてOptuna \cite{akiba2019optuna}のNSGA-II \cite{deb2002nsgaii}を用いている.
  なお, 試行回数はタスクごとに多少異なり, 70,000程度であり, 実行時間は約5時間である.
}%

\begin{figure}[t]
  \centering
  \includegraphics[width=0.95\columnwidth]{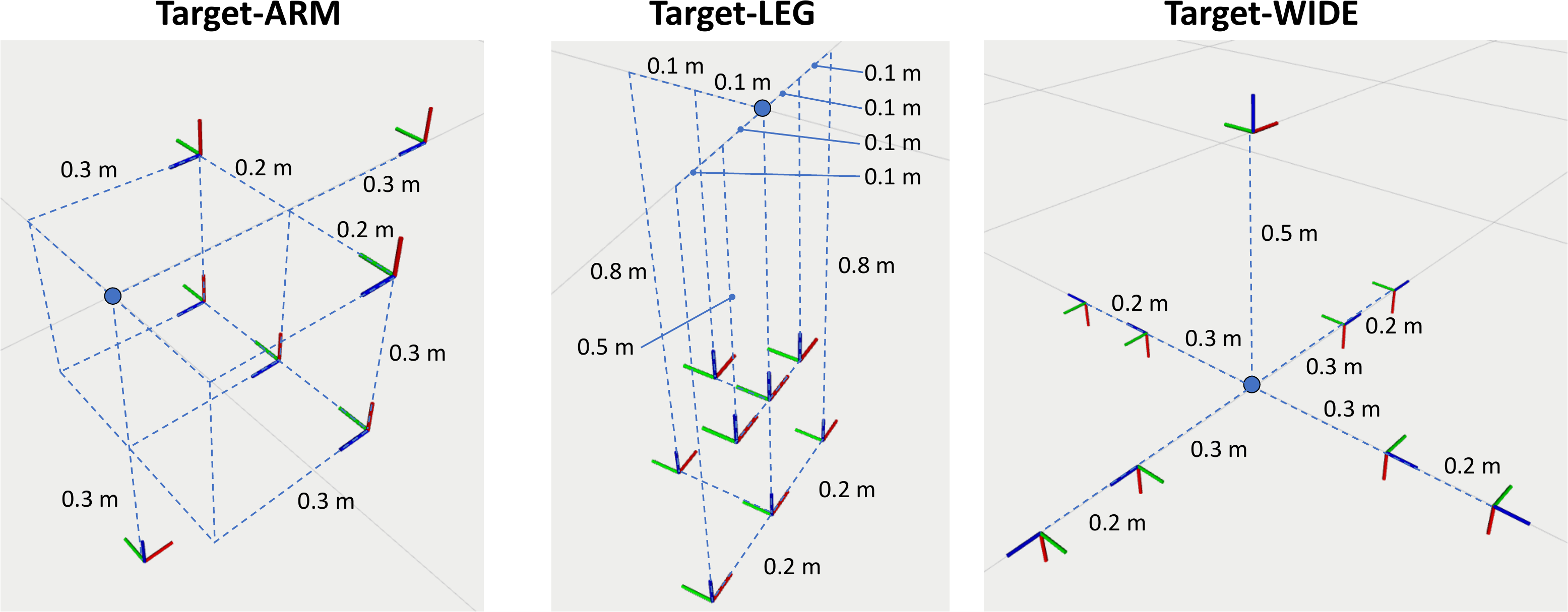}
  \caption{Three types of target positions and orientations: Target-ARM, Target-LEG, and Target-WIDE.}
  \label{figure:target}
\end{figure}

\section{Simulation Experiments} \label{sec:experiment}

\subsection{Experimental Setup}
\switchlanguage%
{%
  In this study, three experiments were conducted while varying the configuration of ($\bm{p}^{ref}_{i}$, $\bm{R}^{ref}_{i}$), as shown in \figref{figure:target}.
  The first experiment, Target-ARM, comprises seven target positions and orientations feasible for human arms.
  The second experiment, Target-LEG, comprises eight target positions and orientations feasible for human legs.
  The third experiment, Target-WIDE, comprises nine target positions and orientations, particularly covering a wider range suitable for industrial robots.
  Note that $N^{jnt}$ was set to 6 for all experiments.
}%
{%
  本研究では, \figref{figure:target}に示すように($\bm{p}^{ref}_{i}$, $\bm{R}^{ref}_{i}$)の設定を変えながら3つの実験を行う.
  1つ目はTarget-ARMであり, 人間の腕が可能なような7つの指令位置姿勢を設定した.
  2つ目はTarget-LEGであり, 人間の脚が可能なような8つの指令位置姿勢を設定した.
  3つ目はTarget-WIDEであり, より広い範囲で, 特に産業用ロボットが可能なような9つの指令位置姿勢を設定した.
  なお, 全ての実験で$N^{jnt}=6$と設定した.
}%

\begin{figure*}[t]
  \centering
  \includegraphics[width=1.6\columnwidth]{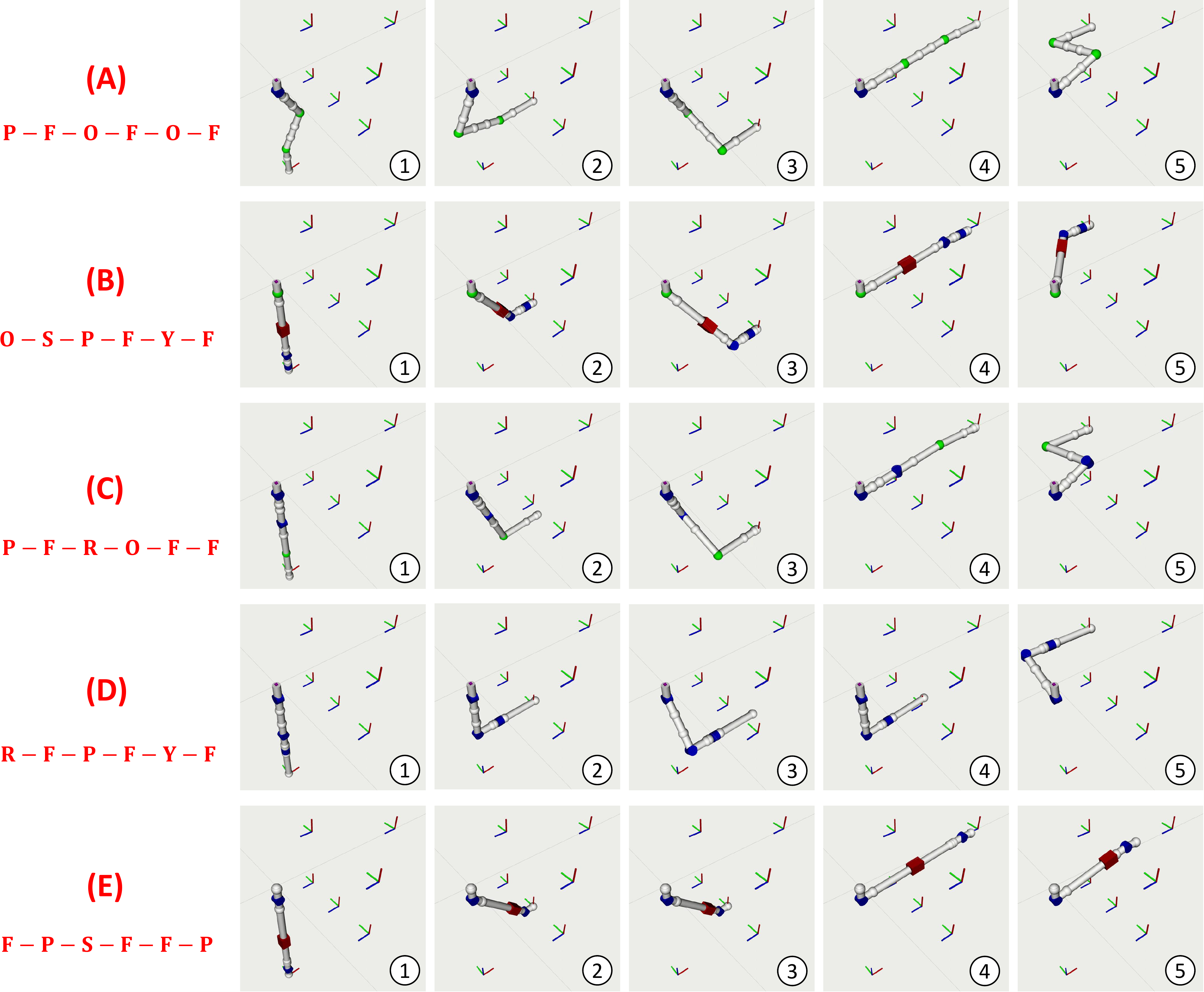}
  \vspace{-1.0ex}
  \caption{Several representative Pareto solutions for Target-ARM. Each snapshot shows the result of inverse kinematics for the target position and orientation.}
  \label{figure:arm-exp}
  \vspace{-1.0ex}
\end{figure*}

\begin{figure}[t]
  \centering
  \includegraphics[width=0.95\columnwidth]{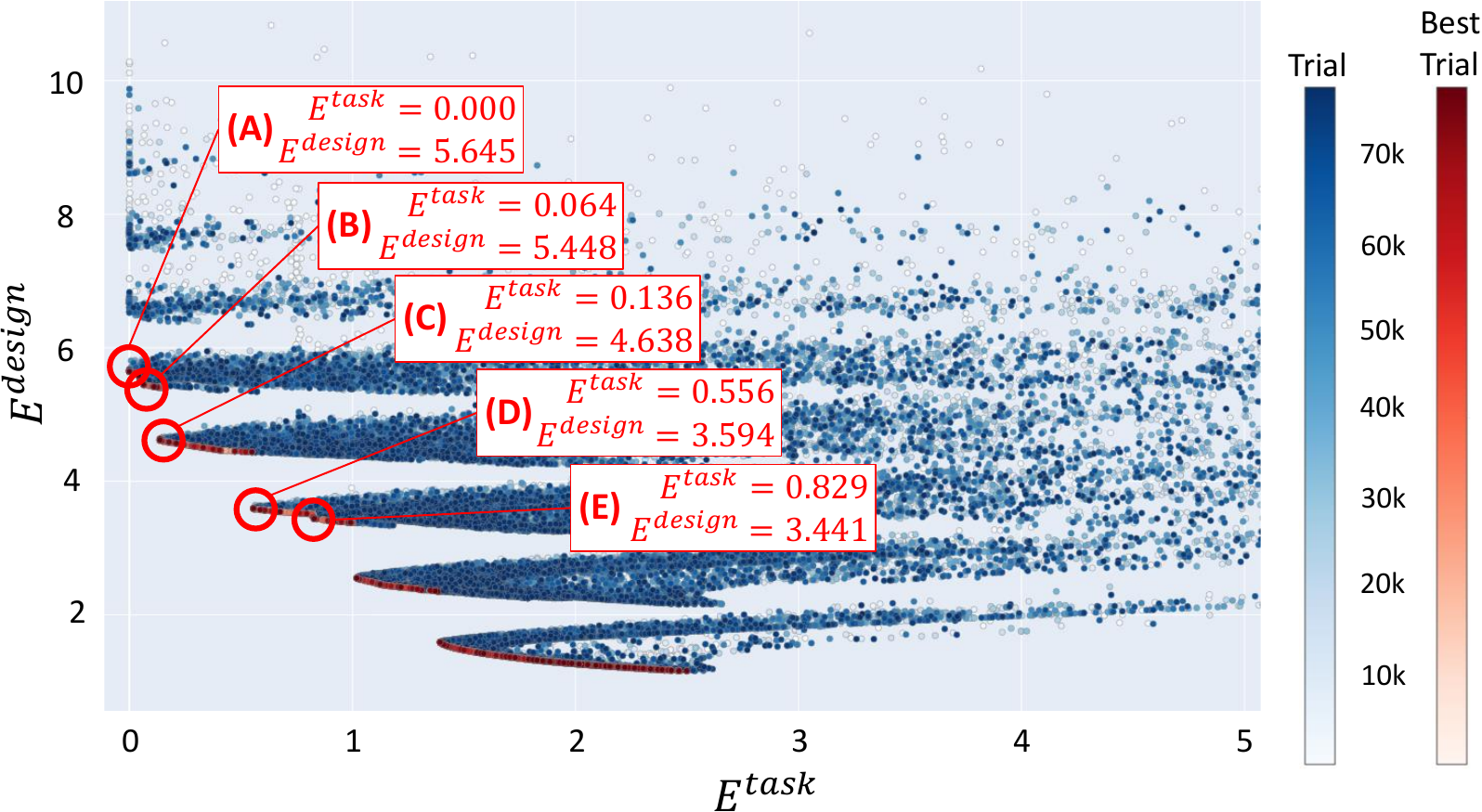}
  \vspace{-1.0ex}
  \caption{Sampling results for Target-ARM. Each point represents a solution, with the Pareto solutions highlighted in red. Solutions (A) to (E) are the representative solutions in \figref{figure:arm-exp}.}
  \label{figure:arm-graph}
  \vspace{-1.0ex}
\end{figure}

\begin{figure*}[t]
  \centering
  \includegraphics[width=1.65\columnwidth]{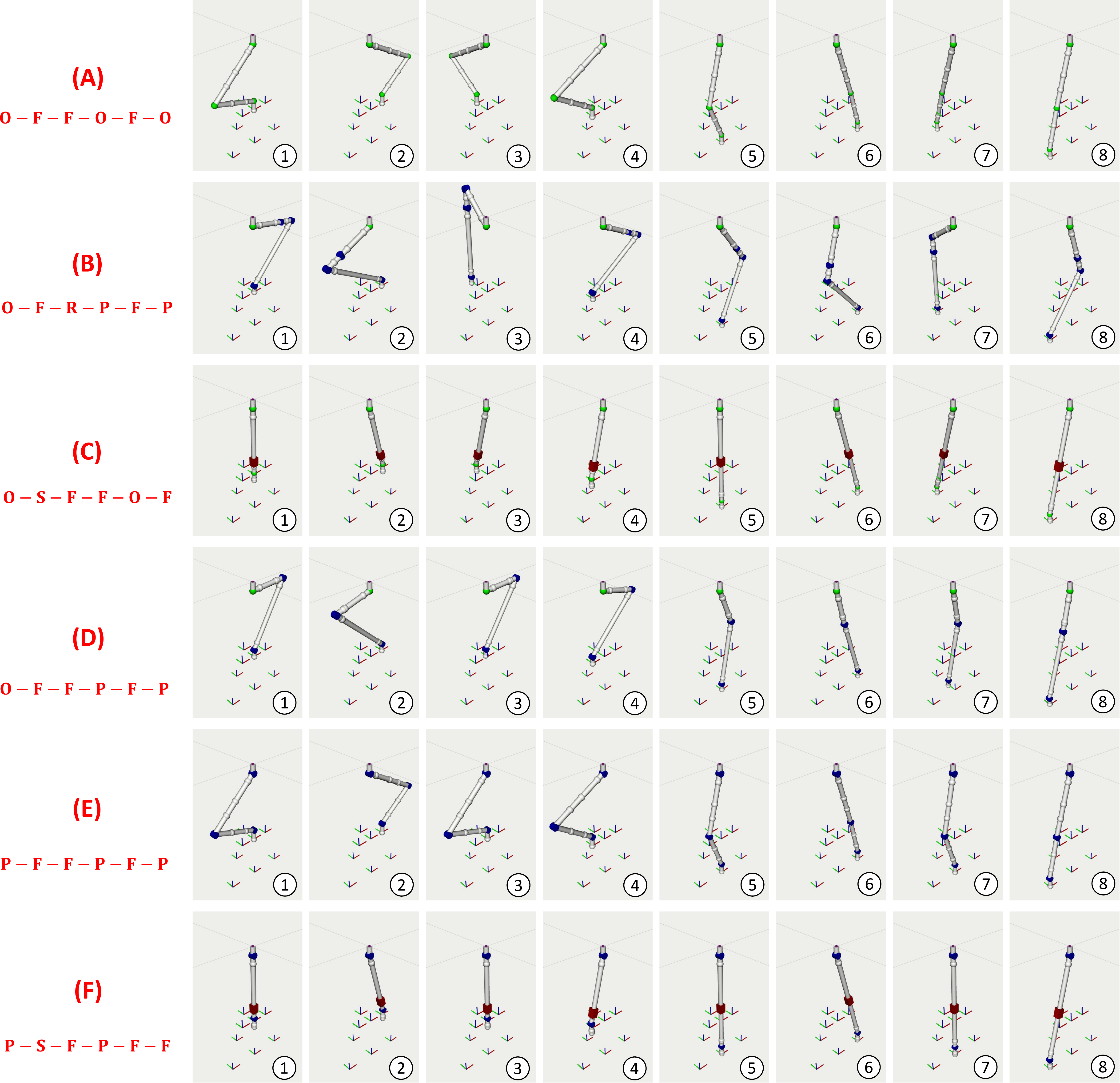}
  \vspace{-1.0ex}
  \caption{Several representative Pareto solutions for Target-LEG. Each snapshot shows the result of inverse kinematics for the target position and orientation.}
  \label{figure:leg-exp}
  \vspace{-1.0ex}
\end{figure*}

\begin{figure}[t]
  \centering
  \includegraphics[width=0.95\columnwidth]{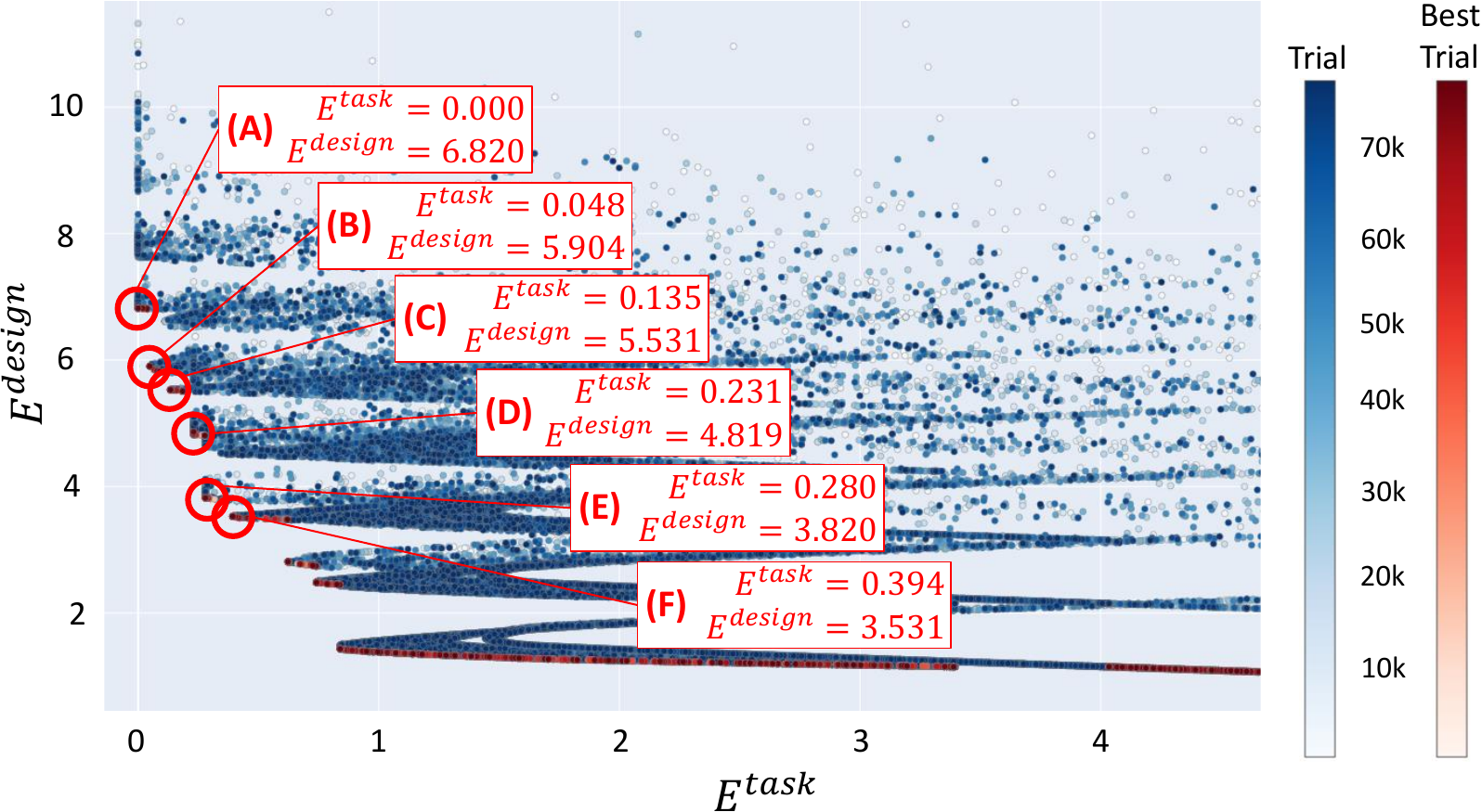}
  \vspace{-1.0ex}
  \caption{Sampling results for Target-LEG. Each point represents a solution, with the Pareto solutions highlighted in red. Solutions (A) to (F) are the representative solutions in \figref{figure:leg-exp}.}
  \label{figure:leg-graph}
  \vspace{-1.0ex}
\end{figure}

\subsection{Results for Target-ARM}
\switchlanguage%
{%
  For Target-ARM, the obtained Pareto solutions are shown in \figref{figure:arm-exp}, and the sampling results are shown in \figref{figure:arm-graph}.

  First, we explain how to interpret the results.
  In \figref{figure:arm-graph}, each point represents a single solution in the sampling process.
  Among these, the red points denote Pareto solutions, indicating superior trade-offs between the horizontal axis $E^{task}$ and the vertical axis $E^{design}$.
  In this study, multiple solutions with different configurations among the Pareto solutions were extracted and presented in \figref{figure:arm-exp}.
  \ctext{1}--\ctext{5} depict the inverse kinematics solutions for each target pose in Target-ARM shown in \figref{figure:target}.
  In the case of symmetrical configurations, some solutions are omitted for clarity.
  Additionally, regarding the values of $E^{design}$, it should be noted that in this study, as the robot's total length $E^{design}_{length}$ never exceeded 1, the integer part of $E^{design}$ represents the degrees of freedom $E^{design}_{joint}$, and the decimal part represents the total length $E^{design}_{length}$.

  The experimental results reveal a diverse set of Pareto solutions.
  (A) and (B) both have 5 DOFs, but (A) has a configuration using only rotational joints, while (B) includes a prismatic joint.
  Both exhibit high performance, with (B) being slightly shorter in total length.
  (C) has 4 DOFs, insufficient to fulfill all target poses, but it achieves high precision sufficiently.
  (C) can be considered a configuration with one fewer degree of freedom from (A).
  (D) and (E) both have 3 DOFs; (D) utilizes only rotational joints, while (E) includes a prismatic joint.
  Unlike (A) or (C), which only include joints for roll and pitch, (D) includes a yaw joint.
  (E) is a configuration with 2 fewer DOFs than (B), moving only in a two-dimensional plane with a prismatic joint.

  When fitting these configurations into existing robots, (A) and (D), despite having different joint orders, resemble a typical humanoid robot arm with shoulder, elbow, and wrist joints, like in HRP-2 \cite{kaneko2004hrp2}.
  Such solutions are reasonable for optimization towards Target-ARM, which emulates the workspace of human arms.
  On the other hand, the structures are not identical; notably, the absence of the yaw joint is a significant difference, driven by the setting that the command values of Target-ARM include almost the same target orientations.
  (C) has a unique configuration with elbow rotation in the roll direction, different from human arms, but it proved effective in this study's settings.
  (B) and (E) are examples of effectively using prismatic joints, resembling a replacement of the human elbow with linear actuators.
  For instance, the surgical robot Da Vinci \cite{freschi2013davinci} has a similar configuration.
  This demonstrates the capability to observe both existing and novel configurations as Pareto solutions simultaneously.
}%
{%
  Target-ARMに対する最適化を行った際に得られたパレート解を\figref{figure:arm-exp}に, サンプリング結果を\figref{figure:arm-graph}に示す.

  まずは結果の見方を示す.
  \figref{figure:arm-graph}は最適化におけるサンプリング結果を示しており, それぞれの点は1つの解を示している.
  このうち, 赤い点はパレート解であり, 横軸$E^{task}$と縦軸$E^{design}$のトレードオフの中でも優れた解である.
  本研究ではパレート解の中でも, 構成の異なる複数の解を取り出し, これを\figref{figure:arm-exp}に示している.
  \ctext{1}--\ctext{5}は\figref{figure:target}のTarget-ARMにおける各指令位置姿勢に対する逆運動学の解を示している.
  なお, 左右対称の場合はいくつかの解を省略して提示している.
  加えて, $E^{design}$の値について, 本研究ではロボットの全長$E^{design}_{length}$が1を超えることが無かったため, $E^{design}$の整数部分が自由度数$E^{design}_{joint}$, 小数点部分が全長$E^{design}_{length}$を表していることに注意されたい.

  実験結果から, 多様なパレート解が得られていることがわかる.
  (A)と(B)はともに5自由度であるが, (A)は回転のみを用いた構成, (B)は直動を含んだ構成である.
  どちらも性能は高いが, (B)の方が多少全長が短くなっている.
  (C)は4自由度であり, 全ての指令位置姿勢を満たすのには不十分であるが, 実現度は非常に高い.
  (C)は(A)から1自由度減らした構成と言える.
  (D)と(E)はともに3自由度であり, (D)は回転のみを用いた構成, (E)は直動を含んだ構成である.
  (D)は(C)から1自由度減らした構成とは言えず, (A)や(C)がロールとピッチの関節のみを含んでいたのに対して, (D)はヨーの関節を含んだ構成である.
  (E)は(B)から2自由度減らした構成であり, 平面のみを動く.

  これを既存のロボットで当てはめると, (A)と(D)は一般的なヒューマノイドロボット\cite{kaneko2004hrp2}のような, 肩・肘・手首を持つアームと似た構成を持つ.
  このような解が出てくることは, 人間の腕のワークスペースと似た設定を行ったTarget-ARMへの最適化としては妥当であると考えられる.
  一方で, もちろん全く同じではなく, 特にTarget-ARMに含まれる点はほとんど同じ指令姿勢を持つことから, ヨー関節が使用されていない点が大きく異なる.
  (C)は肘がロール方向に回転するため人間の腕とは大きく異なっており, あまり見ることのない構成であるが, 本研究の設定には有効であった.
  (B)と(E)は直動関節を上手く使った例であり, 人間の肘部分を直動により置き換えたような構成と言える.
  例えば手術ロボットであるDa Vinci \cite{freschi2013davinci}は似たような構成を持っている.
  このように, 既存の構成, 既存の構成には見られない新しい構成を一度にパレート解として見ることができる.
}%

\begin{figure*}[t]
  \centering
  \includegraphics[width=1.7\columnwidth]{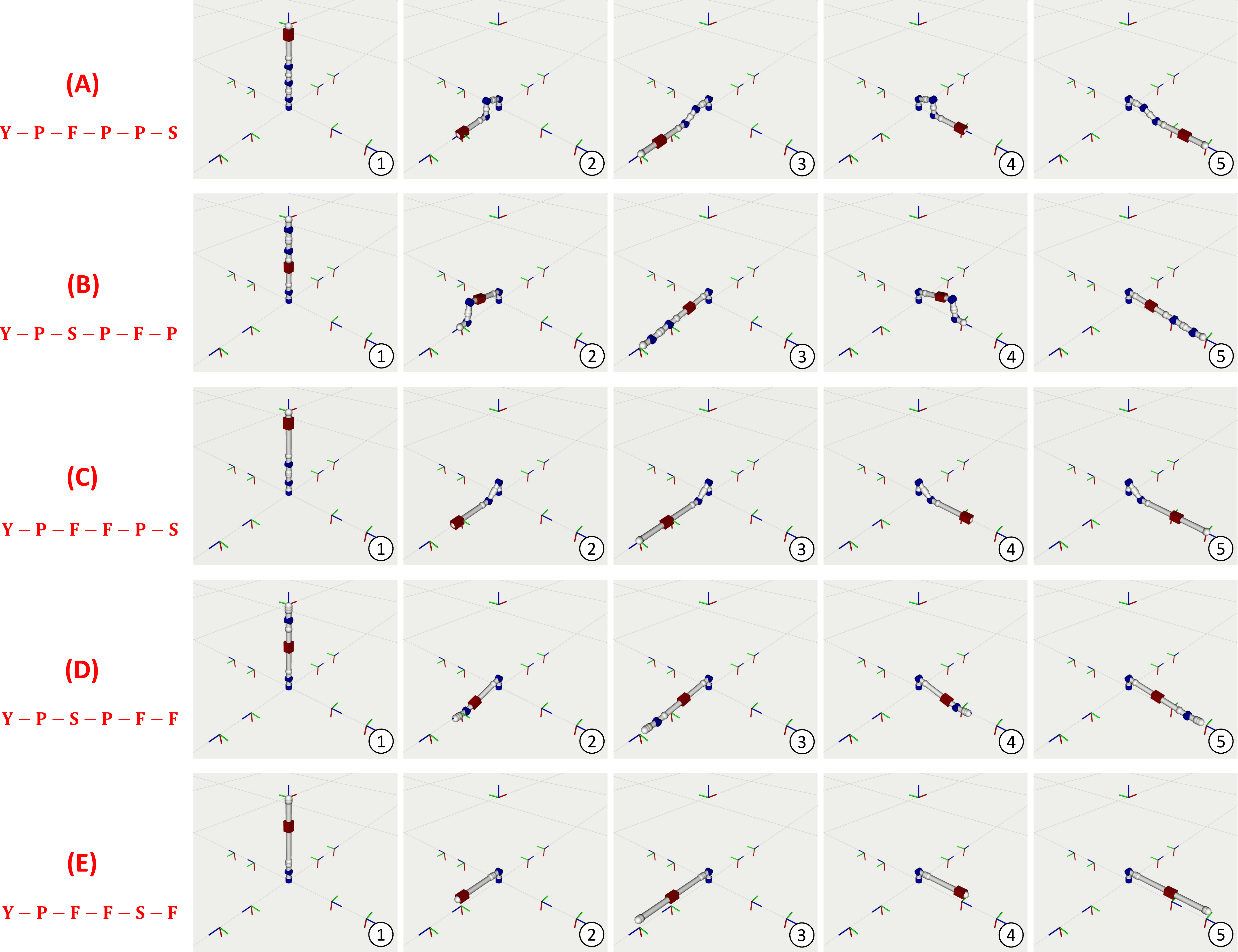}
  \vspace{-1.0ex}
  \caption{Several representative Pareto solutions for Target-WIDE. Each snapshot shows the result of inverse kinematics for the target position and orientation.}
  \label{figure:wide-exp}
  \vspace{-1.0ex}
\end{figure*}

\begin{figure}[t]
  \centering
  \includegraphics[width=0.95\columnwidth]{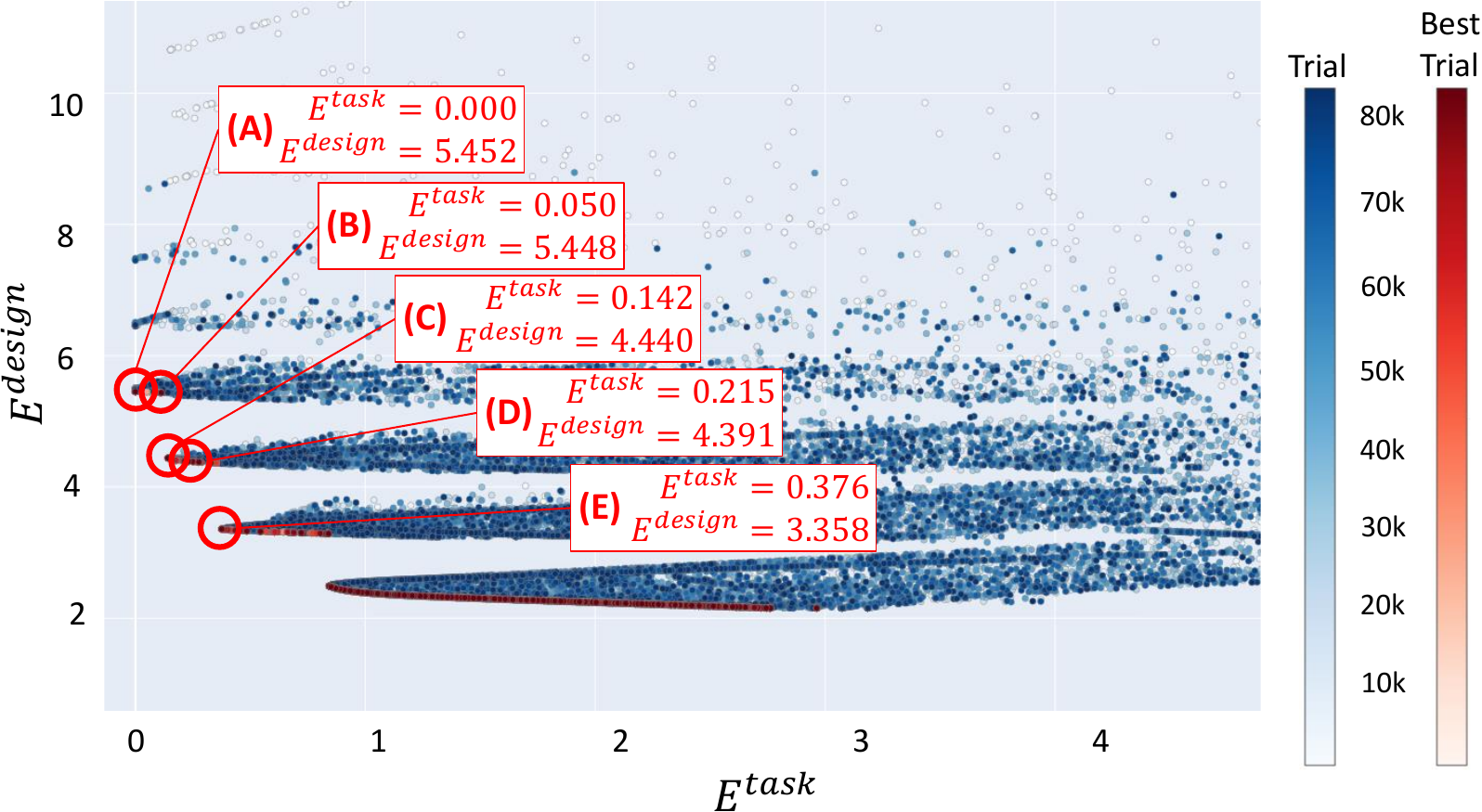}
  \vspace{-1.0ex}
  \caption{Sampling results for Target-WIDE. Each point represents a solution, with the Pareto solutions highlighted in red. Solutions (A) to (E) are the representative solutions in \figref{figure:wide-exp}.}
  \label{figure:wide-graph}
  \vspace{-1.0ex}
\end{figure}

\subsection{Results for Target-LEG}
\switchlanguage%
{%
  For Target-LEG, the obtained Pareto solutions are shown in \figref{figure:leg-exp}, and the sampling results are shown in \figref{figure:leg-graph}.
  (A) has 6 DOFs, featuring a structure composed of three consecutive orthogonal two-axis joints, resulting in high task performance.
  (B) and (C) both have 5 DOFs, with (B) having a configuration similar to (A) but with one fewer degree of freedom, and (C) having a structure with the intermediate orthogonal two-axis joints replaced by a prismatic joint.
  While the overall length of (A) and (B) is approximately 0.9 m, (C) is significantly shorter with a length of 0.531 m.
  (D) has 4 DOFs, representing a configuration with one fewer degree of freedom than (B).
  Finally, (E) and (F) have 3 DOFs each, moving only in a two-dimensional plane.
  (E) can be considered a configuration with one less degree of freedom than (D), and similarly, (F) can be seen as having one less degree of freedom than (C).
  The relationships between (A)-(B)-(C) and (D)-(E)-(F) are similar, as evident from the solution arrangement in \figref{figure:leg-graph}.

  When fitting these configurations into existing robots, (A), (B), (D), and (E) exhibit configurations similar to humanoid robot legs such as HRP-2 \cite{kaneko2004hrp2}, featuring legs with hip, knee, and ankle joints.
  (A) and (E) have thigh and shin links of approximately equal length, while (B) and (D) have much significantly shorter thigh links.
  Furthermore, since all target orientations are aligned, the yaw joint is never used, and the degrees of freedom are constructed solely with roll and pitch. 
  The use of a roll joint in the knee section is a novel and interesting configuration.
  In conventional legs, the knee typically bends in only one direction, but in the setup of this study, this was not replicated.
  (C) and (F) showcase examples of effectively utilizing linear joints, resembling a configuration where the human knee section is replaced by a prismatic joint.
  Similar configurations are observed in robots like SLIDER \cite{wang2020slider}, a humanoid from SCHAFT \cite{urata2016leg}, and a jumping robot RAMIEL \cite{temma2022ramiel}, albeit with different drive mechanisms.
  These robots determine the direction of the link at the root joint, perform vertical movement using a prismatic joint, and control the posture of the foot with ankle joints.
}%
{%
  Target-LEGに対する最適化を行った際に得られたパレート解を\figref{figure:leg-exp}に, サンプリング結果を\figref{figure:leg-graph}に示す.
  実験結果から, 多様なパレート解が得られていることがわかる.
  (A)は6自由度であり, 直交二軸関節を3つ連ねた構造を持ち, 高いタスク達成度を得た.
  (B)と(C)はともに5自由度であり, (B)は(A)から1自由度減らしたような構成, (C)は(A)の中間の直交二軸を直動関節により置き換えたような構成である.
  (A)や(B)の全長が0.9 m程度なのに対して, (C)は0.531 mと全長が圧倒的に短い.
  (D)は4自由度であり, (B)からさらに1自由度減らしたような構成である.
  最後に(E)と(F)は3自由度であり, 平面上のみを動くような構成である.
  (E)は(D)から, (F)は(C)から1自由度減らした構成とも言える.
  (A)-(B)-(C)の関係と, (D)-(E)-(F)の関係は似ており, これは\figref{figure:leg-graph}における解の配置からも読み取れる.

  これを既存のロボットで当てはめると, (A), (B), (D), (E)は比較的一般的なヒューマノイドロボット\cite{kaneko2004hrp2}のような, 股・膝・足首を持つ脚と似た構成を持つ.
  一方で, (A)や(E)は比較的大腿リンクと脛骨リンクが同じような長さであるが, (B)や(D)では大腿リンクがかなり短い.
  また, 指令姿勢が全て同じ方向を向いているため, ヨー関節は一度も使用されておらず, ロールとピッチだけで自由度が構成されている.
  膝の部分にロール関節を利用している点もこれまでにない構成であり興味深い.
  なお, 通常の脚の場合膝は一方方向に曲がらないが, 本研究の設定ではこれは再現できない.
  (C)や(F)は直動関節を上手く使った例であり, 人間の膝部分を直動により置き換えたような構成と言える.
  同様の構成はSLIDER \cite{wang2020slider}やSCHAFTから発表されたロボット \cite{urata2016leg}, 駆動方式は異なるがジャンピングロボット\cite{temma2022ramiel}にも見られる.
  根本の関節で方向を定め, 上下方向の移動を直動関節で行い, 足首関節で姿勢を制御している.
}%

\subsection{Results for Target-WIDE}
\switchlanguage%
{%
  For Target-WIDE, the obtained Pareto solutions are shown in \figref{figure:wide-exp}, and the sampling results are shown in \figref{figure:wide-graph}.
  It can be observed that in all cases (A)-(E), prismatic joints are used, and the two joints from the root link are always arranged in the order of yaw and pitch joints.
  Both (A) and (B) have 5 DOFs each, differing only in the placement of the prismatic joint either at the end or immediately following the yaw and pitch joints.
  Both (C) and (D) have 4 DOFs each, representing configurations with one less degree of freedom compared to (A) and (B), respectively.
  (E) has 3 DOFs, constituting the minimal configuration with yaw, pitch, and prismatic joints.

  When fitting these configurations into existing robots, (B) and (D) exhibit a body configuration with the same order of yaw, pitch, and prismatic joints as the world's first industrial robot, Unimate.
  Moreover, (A) and (C) share a similar structure to a shovel excavator, with a prismatic joint at the end to enhance digging performance.
  Another example with prismatic joints is the Mujin TuckBot \cite{mujin2023tuckbot}, where due to its small workspace, the prismatic joint is placed closest to the base, followed by yaw and pitch joints.
  This contrasts with the configuration seen in setups with a wide workspace like Target-WIDE.
  Additionally, Fetch \cite{fetch2015fetch} employs a prismatic joint extending in the $z$ direction from the root link, while a three-axis NC machine has a prismatic joint extending in the $xy$ direction from the root link.
  The former is confirmed to occur depending on the setting of the target positions, while the latter cannot meet the constraint of having the links aligned in a straight line in the initial state, so was not observed in the context of this study.
}%
{%
  Target-WIDEに対する最適化を行った際に得られたパレート解を\figref{figure:wide-exp}に, サンプリング結果を\figref{figure:wide-graph}に示す.
  (A)-(E)の全てで直動関節が用いられており, かつ根本から2関節は必ずヨー関節, ピッチ関節の順に並んでいることが分かる.
  (A)と(B)はともに5自由度であり, 直動関節が末端にあるか, ヨー関節, ピッチ関節の次にあるかの違いである.
  (C)と(D)はともに4自由度であり, それぞれ(A)と(B)から1自由度減らした構成と言える.
  (E)は3自由度であり, ヨー, ピッチ, 直動という最小構成である.

  これを既存のロボットで当てはめると, (B)と(D)は世界初の産業用ロボットであるUnimateと同様な, ヨー, ピッチ, 直動という順番の身体構成を持つことがわかる.
  また, (A)や(C)は, 直動関節を末端に持ち掘削性能を上げたショベルカーと同様の構造を持っている.
  この他にも直動関節を使った例としてはMujin TuckBot \cite{mujin2023tuckbot}があるが, これはワークスペースが小さいため直動関節が最も手前に配置されており, その先にヨー, ピッチと続く.
  Target-WIDEのような広いワークスペースの設定では見ることの構成であろう.
  加えてFetch \cite{fetch2015fetch}はルートリンクから$z$方向に直動が, 3軸NCマシンはルートリンクから$xy$方向に直動が伸びる.
  前者は指令位置の設定次第で発現することを確認しているが, 後者は初期状態で一直線にリンクが並ぶという制約を満たせないため, 本研究の設定では見られない.
}%

\section{Conclusion and Future Works} \label{sec:conclusion}
\switchlanguage%
{%
  In this study, we explored diverse body configurations in modular robots encompassing both rotational and prismatic joints by utilizing black-box multi-objective optimization.
  Defining each joint module as a separate Xacro file, we represent various body configurations determined by the arrangement of these joint modules and lengths of the links.
  By incorporating not only task achievement but also metrics related to the robot's body design into the objective functions, we are able to obtain Pareto solutions of robot configurations with diverse bodies, rather than a single solution.
  From the experiments, we obtained not only well-established structures commonly found in humanoids and industrial robots but also rare and unconventional body structures.
  The results from task achievement and body design metrics related to simple kinematics alone yielded highly intriguing findings.
  This study provides a foundation for exploring new body configurations while considering existing ones in robots with diverse joints.

  There are numerous potential extensions for this research.
  Firstly, further exploration is needed for bodies with a wider variety of joints, including closed-links, wire-driven structures, wheels, and other diverse joint types, in addition to rotational and prismatic joints.
  Configurations with branching, giving rise to limbs similar to humanoid structures, are also intriguing \cite{zhao2020robogrammar}.
  To combine more complex structures while maintaining consistency, appropriate constraint conditions must be established, posing a significant challenge for future research.
  Similarly, there is a need to advance the definition of a more diverse set of objective functions.
  Evaluating whether existing robot structures are suitable, exploring new structures for various tasks, and experimenting with different objective function designs are crucial aspects.
  Considering the importance of dynamics, combining control methods such as model predictive control and reinforcement learning is necessary for practical robot structure exploration.
  Although solutions for several of these challenges already exist, we aim to focus on the autonomous emergence of diverse bodies by combining multiple joint structures in further research.
}%
{%
  本研究では, 回転関節と直動関節を含むモジュラー型ロボットにおける多目的ブラック最適化を用いた多様な身体構成の探索を行った.
  各関節モジュールを個別のXacroファイルとして定義し, どの関節をどの長さで接続するかによって多様な身体構成を表現した.
  目的関数にはタスクの達成度だけでなく, ロボットの身体設計に関する指標も含めることで, たった一つの解ではなく, 多様な身体を持つロボット構成のパレート解を得ることができた.
  実験から, 既存のヒューマノイドや産業用ロボットに見られるような実績ある構造だけでなく, あまり見ることのない珍しい身体構造も得ることができた.
  単純なキネマティクスに関するタスク達成度と身体設計指標だけでも, 非常に興味深い結果が得られている.
  本研究により, 多様な関節を含むロボットに置いて, 既存の身体構成も考慮に入れつつ, 新たな身体構成を探索するための一つの基盤が構築できたと考えている.

  一方で, 本研究にはさらなる拡張が多数考えられる.
  まず, より多様な関節を含む身体の探索を進めなければならない.
  直動と回転関節だけでなく, 閉リンクやワイヤ駆動構造, 車輪など, 他の様々な関節も含めた探索が必要である.
  シリアルなリンク構造だけでなく, 途中で枝分かれを起こし, ヒューマノイドのように四肢が生まれる設定も興味深い\cite{zhao2020robogrammar}.
  それらより複雑な構造を整合性を保ちながら組み合わせるためには, 適切な制約条件を設定する必要があり, 今後の大きな課題となっている.
  同様に, より多様な目的関数の設定も進める必要がある.
  これまで作られてきた既存のロボットの構造は適切なのか, より適切な新しい構造は無いのか, より多様なタスクについて, 様々な目的関数設計を試したい.
  ダイナミクスの考慮も重要なため, モデル予測制御や強化学習などの制御手法と組み合わせも, より実用的なロボット構造の探索には必要だろう.
  これらの課題のうちいくつかは既に解法があるが, 多様な身体の自律的な発現という観点から, 複数の関節構造の組み合わせに着目して研究を進めていきたい.
}%

{
  \bibliographystyle{IEEEtran}
  \bibliography{main}
}

\end{document}